# AND/OR Importance Sampling

**Vibhav Gogate and Rina Dechter**
Department of Information and Computer Science,
University of California, Irvine, CA 92697,
{vgogate,dechter}@ics.uci.edu

## Abstract

The paper introduces AND/OR importance sampling for probabilistic graphical models. In contrast to importance sampling, AND/OR importance sampling caches samples in the AND/OR space and then extracts a new sample mean from the stored samples. We prove that AND/OR importance sampling may have lower variance than importance sampling; thereby providing a theoretical justification for preferring it over importance sampling. Our empirical evaluation demonstrates that AND/OR importance sampling is far more accurate than importance sampling in many cases.

## 1 Introduction

Many problems in graphical models such as computing the probability of evidence in Bayesian networks, solution counting in constraint networks and computing the partition function in Markov random fields are *summation problems*, defined as a sum of a function over a domain. Because these problems are NP-hard, sampling based techniques are often used to approximate the sum. The focus of the current paper is on *importance sampling*.

The main idea in importance sampling [Geweke, 1989, Rubinstein, 1981] is to transform the summation problem to that of computing a weighted average over the domain by using a special distribution called the proposal (or importance) distribution. Importance sampling then generates samples from the proposal distribution and approximates the true average over the domain by an average over the samples; often referred to as the sample average. The sample average is simply a ratio of the sum of sample weights and the number of samples, and it can be computed in a *memory-less* fashion since it requires keeping only these two quantities in memory.

The main idea in this paper is to equip importance sampling with memoization or caching in order to exploit conditional independencies that exist in the graphical model. Specifically, we cache the samples on an AND/OR tree or graph [Dechter and Mateescu, 2007] which respects the structure of the graphical model and then compute a new weighted average over that AND/OR structure, yielding, as we show, an unbiased estimator that has a smaller variance than the importance sampling estimator. Similar to AND/OR search [Dechter and Mateescu, 2007], our new AND/OR importance sampling scheme recursively combines samples that are cached in independent components yielding an increase in the effective sample size which is part of the reason that its estimates have lower variance.

We present a detailed experimental evaluation comparing importance sampling with AND/OR importance sampling on Bayesian network benchmarks. We observe that the latter outperforms the former on most benchmarks and in some cases quite significantly.

The rest of the paper is organized as follows. In the next section, we describe preliminaries on graphical models, importance sampling and AND/OR search spaces. In sections 3, 4 and 5 we formally describe AND/OR importance sampling and prove that its sample mean has lower variance than conventional importance sampling. Experimental results are described in section 6 and we conclude with a discussion of related work and summary in section 7.

## 2 Preliminaries

We represent sets by bold capital letters and members of a set by capital letters. An assignment of a value to a variable is denoted by a small letter while bold small letters indicate an assignment to a set of variables.

**Definition 2.1 (belief networks).** A *belief network (BN)* is a graphical model $\mathscr{R} = (\mathbf{X}, \mathbf{D}, \mathbf{P})$, where $\mathbf{X} = \{X_1, \ldots, X_n\}$ is a set of random variables over multi-valued domains $\mathbf{D} = \{\mathbf{D_1}, \ldots, \mathbf{D_n}\}$. Given a directed acyclic graph $G$ over $\mathbf{X}$, $\mathbf{P} = \{P_i\}$, where $P_i = P(X_i|\mathbf{pa}(X_i))$ are conditional probability tables (CPTs) associated with each $X_i$. $\mathbf{pa}(X_i)$ is the set of parents of the variable $X_i$ in $G$. A belief network represents a probability distribution over $\mathbf{X}$, $P(\mathbf{X}) =$ $\prod_{i=1}^{n} P(X_i|\mathbf{pa}(X_i))$. An *evidence set* $\mathbf{E}=\mathbf{e}$ is an instantiated subset of variables.The moral graph (or primal graph) of a belief network is the undirected graph obtained by connecting the parent nodes and removing direction.

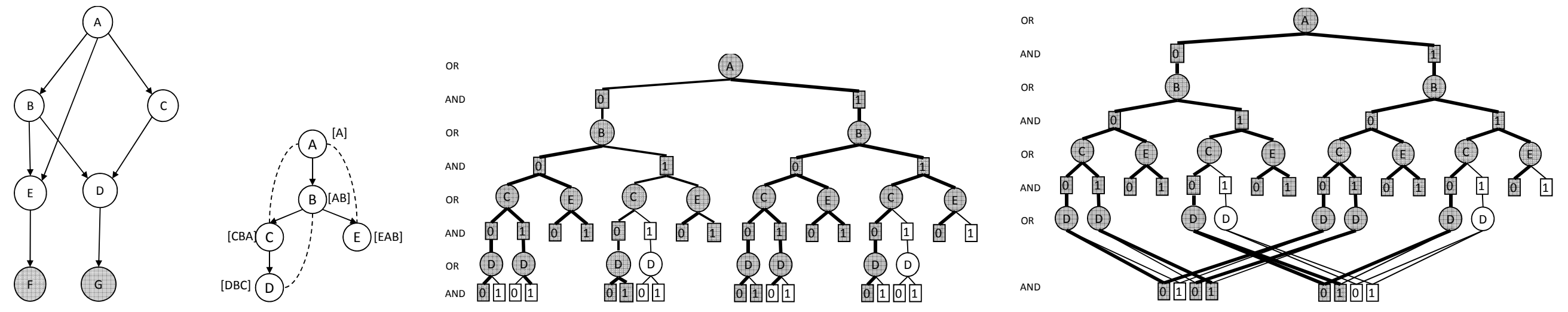

(a) (b) (c) (d)

Figure 1: (a) Bayesian Network, (b) Pseudo-tree (c) AND/OR tree (d) AND/OR search graph

**Definition 2.2 (Probability of Evidence).** Given a belief network $\mathscr{R}$ and evidence $\mathbf{E}=\mathbf{e}$, the probability of evidence $P(\mathbf{E}=\mathbf{e})$ is defined as:

$$P(\mathbf{e}) = \sum_{\mathbf{X}\setminus\mathbf{E}} \prod_{j=1}^{n} P(X_j|\mathbf{pa}(X_j))_{|\mathbf{E}=\mathbf{e}} \quad (1)$$

The notation $h(\mathbf{X})_{|\mathbf{E}=\mathbf{e}}$ stands for a function $h$ over $\mathbf{X}\setminus\mathbf{E}$ with the assignment $\mathbf{E}=\mathbf{e}$.

### 2.1 AND/OR search spaces

We can compute probability of evidence by search, by accumulating probabilities over the search space of instantiated variables. In the simplest case, this process defines an OR search tree, whose nodes represent partial variable assignments. This search space does not capture the structure of the underlying graphical model. To remedy this problem, [Dechter and Mateescu, 2007] introduced the notion of AND/OR search space. Given a bayesian network $\mathscr{R}=(\mathbf{X},\mathbf{D},\mathbf{P})$, its AND/OR search space is driven by a pseudo tree defined below.

**Definition 2.3 (Pseudo Tree).** Given an undirected graph $G=(V,E)$, a directed rooted tree $T=(V,E)$ defined on all its nodes is called pseudo tree if any arc of G which is not included in $E$ is a back-arc, namely it connects a node to an ancestor in $T$.

**Definition 2.4 (Labeled AND/OR tree).** Given a graphical model $\mathscr{R}=\langle\mathbf{X},\mathbf{D},\mathbf{P}\rangle$, its primal graph $G$ and a backbone pseudo tree $T$ of $G$, the associated AND/OR search tree, has alternating levels of AND and OR nodes. The OR nodes are labeled $X_i$ and correspond to the variables. The AND nodes are labeled $\langle X_i, x_i\rangle$ and correspond to the value assignments in the domains of the variables. The structure of the AND/OR search tree is based on the underlying backbone tree $T$. The root of the AND/OR search tree is an OR node labeled by the root of $T$.

Each OR arc, emanating from an OR node to an AND node is associated with a **label** which can be derived from the CPTs of the bayesian network [Dechter and Mateescu, 2007]. Each OR node and AND node is also associated with a **value** that is used for computing the quantity of interest.

Semantically, the OR states represent alternative assignments, whereas the AND states represent problem decomposition into independent subproblems, all of which need be solved. When the pseudo-tree is a chain, the AND/OR search tree coincides with the regular OR search tree. The probability of evidence can be computed from a labeled AND/OR tree by recursively computing the value of all nodes from leaves to the root [Dechter and Mateescu, 2007].

**Example 2.5.** Figure 1(a) shows a bayesian network over seven variables with domains of $\{0,1\}$. $F$ and $G$ are evidence nodes. Figure 1(c) shows the AND/OR-search tree for the bayesian network based on the Pseudo-tree in Figure 1(b). Note that because $F$ and $G$ are instantiated, the search space has only 5 variables.

### 2.2 Computing Probability of Evidence Using Importance Sampling

Importance sampling [Rubinstein, 1981] is a simulation technique commonly used to evaluate the sum, $M=\sum_{\mathbf{x}\in\mathbf{X}} f(\mathbf{x})$ for some real function $f$. The idea is to generate samples $\mathbf{x^1},\ldots,\mathbf{x^N}$ from a proposal distribution $Q$ (satisfying $f(\mathbf{x})>0 \Rightarrow Q(\mathbf{x})>0$) and then estimate $M$ as follows:

$$M = \sum_{\mathbf{x}\in\mathbf{X}} f(\mathbf{x}) = \sum_{\mathbf{x}\in\mathbf{X}} \frac{f(\mathbf{x})}{Q(\mathbf{x})} Q(\mathbf{x}) = \mathbb{E}_Q[\frac{f(\mathbf{x})}{Q(\mathbf{x})}] \quad (2)$$

$$\widehat{M} = \frac{1}{N}\sum_{i=1}^{N} w(\mathbf{x}^i)\ ,\ where\ \ w(\mathbf{x}^i) = \frac{f(\mathbf{x}^i)}{Q(\mathbf{x}^i)} \quad (3)$$

$w$ is often referred to as the sample weight. It is known that the expected value $\mathbb{E}(\widehat{M})=M$ [Rubinstein, 1981].

To compute the probability of evidence by importance sampling, we use the substitution:

$$f(\mathbf{x}) = \prod_{j=1}^{n} P(X_j|\mathbf{pa}(X_j))_{|\mathbf{E}=\mathbf{e}} \quad (4)$$

Several choices are available for the proposal distribution $Q(\mathbf{x})$ ranging from the prior distribution as in likelihood weighting to more sophisticated alternatives such as IJGP-Sampling [Gogate and Dechter, 2005] and EPIS-BN [Yuan and Druzdzel, 2006] where the output of belief propagation is used to compute the proposal distribution.

As in prior work [Cheng and Druzdzel, 2000], we assume that the proposal distribution is expressed in a factored product form: $Q(\mathbf{X}) = \prod_{i=1}^{n} Q_i(X_i|X_1,\ldots,X_{i-1}) = \prod_{i=1}^{n} Q_i(X_i|\mathbf{Y_i})$, where $\mathbf{Y_i} \subseteq \{X_1,\ldots,X_{i-1}\}$, $Q_i(X_i|\mathbf{Y_i}) = Q(X_i|X_1,\ldots,X_{i-1})$ and $|\mathbf{Y_i}| < c$ for some constant $c$. We can generate a full sample from $Q$ as follows. For $i = 1$ to $n$, sample $X_i = x_i$ from the conditional distribution $Q(X_i|X_1 = x_1,\ldots,X_{i-1} = x_{i-1})$ and set $X_i = x_i$.

# 3 AND/OR importance sampling

We first discuss computing expectation by parts; which forms the backbone of AND/OR importance sampling. We then present the AND/OR importance sampling scheme formally and derive its properties.

## 3.1 Estimating Expectation by Parts

In Equation 2, the expectation of a multi-variable function is computed by summing over the entire domain. This method is clearly inefficient because it does not take into account the decomposition of the multi-variable function as we illustrate below.

Consider the tree graphical model given in Figure 2(a). Let $A = a$ and $B = b$ be the evidence variables. Let $Q(ZXY) = Q(Z)Q(X|Z)Q(Y|Z)$ be the proposal distribution. For simplicity, let us assume that $f(Z) = P(Z)$, $f(XZ) = P(Z|X)P(A = a|X)$ and $f(YZ) = P(Z|Y)P(B = b|Y)$. We can express probability of evidence $P(a,b)$ as:

$$\begin{aligned} P(a,b) &= \sum_{XYZ} \frac{f(Z)f(XZ)f(YZ)}{Q(Z)Q(X|Z)Q(Y|Z)} Q(Z)Q(X|Z)Q(Y|Z) \\ &= \mathbb{E}\left[\frac{f(Z)f(XZ)f(YZ)}{Q(Z)Q(X|Z)Q(Y|Z)}\right] \end{aligned} \quad (5)$$

We can decompose the expectation in Equation 5 into smaller components as follows:

$$P(a,b) = \sum_{Z} \frac{f(Z)Q(Z)}{Q(Z)} \left(\sum_{X} \frac{f(XZ)Q(X|Z)}{Q(X|Z)}\right)\left(\sum_{Y} \frac{f(YZ)Q(Y|Z)}{Q(Y|Z)}\right) \quad (6)$$

The quantities in the two brackets in Equation 6 are, by definition, conditional expectations of a function over $X$ and $Y$ respectively given $Z$. Therefore, Equation 6 can be written as:

$$P(a,b) = \sum_{Z} \frac{f(Z)}{Q(Z)} \mathbb{E}\left[\frac{f(XZ)}{Q(X|Z)}|Z\right] \mathbb{E}\left[\frac{f(YZ)}{Q(Y|Z)}|Z\right] Q(Z) \quad (7)$$

By definition, Equation 7 can be written as:

$$P(a,b) = \mathbb{E}\left[\frac{f(Z)}{Q(Z)} \mathbb{E}\left[\frac{f(XZ)}{Q(X|Z)}|Z\right] \mathbb{E}\left[\frac{f(YZ)}{Q(Y|Z)}|Z\right]\right] \quad (8)$$

We will refer to Equations of the form 8 as *expectation by parts* borrowing from similar terms such as integration and summation by parts. If the domain size of all variables is $d = 3$, for example, computing expectation using Equation 5 would require summing over $d^3 = 3^3 = 27$ terms while computing the same expectation by parts would require summing over $d + d^2 + d^2 = 3 + 3^2 + 3^2 = 21$ terms. Therefore, exactly computing expectation by parts is clearly more efficient.

Importance sampling ignores the decomposition of expectation while approximating it by the sample average. Our new algorithm estimates the true expectation by decomposing it into several conditional expectations and then approximating each by an appropriate weighted average over the samples. Since computing expectation by parts is less complex than computing expectation by summing over the domain; we expect that approximating it by parts will be easier as well. We next illustrate how to estimate expectation by parts on our example Bayesian network given in Figure 2(a).

Assume that we are given samples $(z^1,x^1,y^1),\ldots,(z^N,x^N,y^N)$ generated from $Q$ decomposed according to Figure 2(a). For simplicity, let $\{0,1\}$ be the domain of $Z$ and let $Z = 0$ and $Z = 1$ be sampled $N_0$ and $N_1$ times respectively. We can approximate $\mathbb{E}\left[\frac{f(XZ)}{Q(X|Z)}|Z\right]$ and $\mathbb{E}\left[\frac{f(YZ)}{Q(Y|Z)}|Z\right]$ by $g_X\widehat{(Z = j)}$ and $g_Y\widehat{(Z = j)}$ defined below:

$$\begin{aligned} g_X\widehat{(Z = j)} &= \frac{1}{N_j}\sum_{i=1}^{N} \frac{f(x^i, Z = j)I(x^i, Z = j)}{Q(x^i, Z = j)} \\ g_Y\widehat{(Z = j)} &= \frac{1}{N_j}\sum_{i=1}^{N} \frac{f(y^i, Z = j)I(y^i, Z = j)}{Q(y^i, Z = j)} \end{aligned} \quad (9)$$

where $I(x^i, Z = j)$ (or $I(y^i, Z = j)$) is an indicator function which is 1 iff the tuple $(x^i, Z = j)$ ( or $(y^i, Z = j)$ ) is generated in any of the $N$ samples and 0 otherwise.

From Equation 8, we can now derive the following unbiased estimator for $P(a,b)$:

$$\widehat{P(a,b)} = \frac{1}{N}\sum_{j=0}^{1} \frac{N_j f(Z = j) g_X\widehat{(Z = j)} g_Y\widehat{(Z = j)}}{Q(Z = j)} \quad (10)$$

Importance sampling on the other hand would estimate $P(a,b)$ as follows:

$$\begin{aligned} \widetilde{P(a,b)} = &\frac{1}{N}\sum_{j=0}^{1} N_j \frac{f(Z = j)}{Q(Z = j)} \\ &\times \frac{1}{N_j}\sum_{i=1}^{N} \frac{f(x^i, Z = j)f(y^i, Z = j)}{Q(x^i|Z = j)Q(Y^i|Z = j)} I(x^i, y^i, Z = j) \end{aligned} \quad (11)$$

where $I(x^i, y^i, Z = j)$ is an indicator function which is 1 iff the tuple $(x^i, y^i, Z = j)$ is generated in any of the $N$ samples and 0 otherwise.

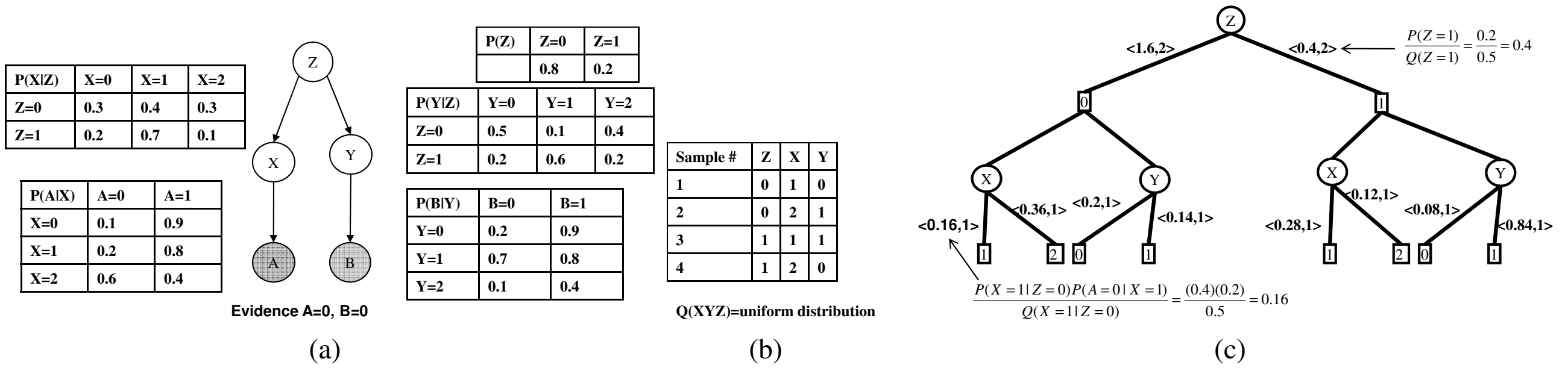

Figure 2: (a) Bayesian Network, its CPTs, (b) Proposal Distribution and Samples (c) AND/OR sample tree

Equation 10 which is an unbiased estimator of expectation by parts given in Equation 8 provides another rationale for preferring it over the usual importance sampling estimator given by Equation 11. In particular in Equation 10, we estimate two functions defined over the random variables $X|Z=z$ and $Y|Z=z$ respectively from the generated samples. In importance sampling, on the other hand, we estimate a function over the joint random variable $XY|Z=z$ using the generated samples. Because the samples for $X|Z=z$ and $Y|Z=z$ are considered independently in Equation 10, $N_j$ samples drawn over the joint random variable $XY|Z=z$ in Equation 11 correspond to a larger set $N_j * N_j = N_j^2$ of virtual samples. We know that [Rubinstein, 1981] the variance (and therefore the mean-squared error) of an unbiased estimator decreases with an increase in the effective sample size. Consequently, our new estimation technique will have lower error than the conventional approach.

In the following subsection, we discuss how the AND/OR structure can be used for estimating expectation by parts yielding the *AND/OR importance sampling* scheme.

### 3.2 Computing Sample Mean in AND/OR-space

In this subsection, we formalize the ideas of estimating expectation by parts on a general AND/OR tree starting with some required definitions. We define the notion of an AND/OR sample tree which is restricted to the generated samples and which will be used to compute the AND/OR sample mean. The labels on this AND/OR tree are set to account for the importance weights.

**Definition 3.1 (Arc Labeled AND/OR Sample Tree).** Given a a graphical model $\mathscr{R} = \langle \mathbf{X}, \mathbf{D}, \mathbf{P} \rangle$, a pseudo-tree $T(V,E)$ , a proposal distribution $Q = \prod_{i=1}^{n} Q(X_i|\mathbf{Anc}(X_i))$ such that $\mathbf{Anc(X_i)}$ is a subset of all ancestors of $X_i$ in $T$, a sequence of assignments (samples) $\mathbf{S}$ and a complete AND/OR search tree $\phi_T$, an AND/OR sample tree $S_{AOT}$ is constructed from $\phi_T$ by removing all edges and corresponding nodes which are not in $\mathbf{S}$ i.e. they are not sampled.

The Arc-label for an OR node $X_i$ to an AND node $X_i = x_i$ in $S_{AOT}$ is a pair $\langle w, \# \rangle$ where:

- $w = \frac{P(X_i=x_i,\mathbf{anc}(x_i))}{Q(X_i=x_i|\mathbf{anc}(x_i))}$ is called the weight of the arc. $\mathbf{anc}(x_i)$ is the assignment of values to all variables from the node $X_i$ to the root node of $S_{AO}$ and $P(X_i = x_i, \mathbf{anc}(x_i))$ is the product of all functions in $\mathscr{R}$ that mention $X_i$ but do not mention any variable ordered below it in $T$ given $(X_i = x_i, \mathbf{anc}(x_i))$.
- # is the frequency of the arc. Namely, it is equal to the number of times the assignment $(X_i = x_i, \mathbf{anc}(x_i))$ is sampled.

**Example 3.2.** Consider again the Bayesian network given in Figure 2(a). Assume that the proposal distribution $Q(XYZ)$ is uniform. Figure 2(b) shows four hypothetical random samples drawn from $Q$. Figure 2(c) shows the AND/OR sample tree over the four samples. Each arc from an OR node to an AND node in the AND/OR sample tree is labeled with appropriate frequencies and weights according to Definition 3.1. Figure 2(c) shows the derivation of arc-weights for two arcs.

The main virtue of arranging the samples on an AND/OR sample tree is that we can exploit the independencies to define the *AND/OR sample mean*.

**Definition 3.3 (AND/OR Sample Mean).** Given a AND/OR sample tree with arcs labeled according to Definition 3.1, the **value** of a node is defined recursively as follows. The value of leaf AND nodes is ”1” and the value of leaf OR nodes is ”0”. Let $\mathbf{C(n)}$ denote the child nodes and $v(n)$ denotes the value of node $n$. If $n$ is a AND node then: $v(n) = \prod_{n' \in \mathbf{C(n)}} v(n')$ and if $n$ is a OR node then

$$v(n) = \frac{\sum_{n' \in \mathbf{C(n)}} (\#(n,n')w(n,n')v(n'))}{\sum_{n' \in \mathbf{C(n)}} \#(n,n')}$$

The **AND/OR sample mean** is the value of the root node.

We can show that the value of an OR node is equal to an unbiased estimate of the conditional expectation of the variable at the OR node given an assignment from the root to the parent of the OR node. Since all variables, except the evidence variables are unassigned at the root node, the value of the root node equals the AND/OR sample mean which is an unbiased estimate of probability of evidence. Formally,

THEOREM 3.4. *The AND/OR sample mean is an unbiased estimate of probability of evidence.*

**Example 3.5.** The calculations involved in computing the sample mean on the AND/OR sample tree on our example

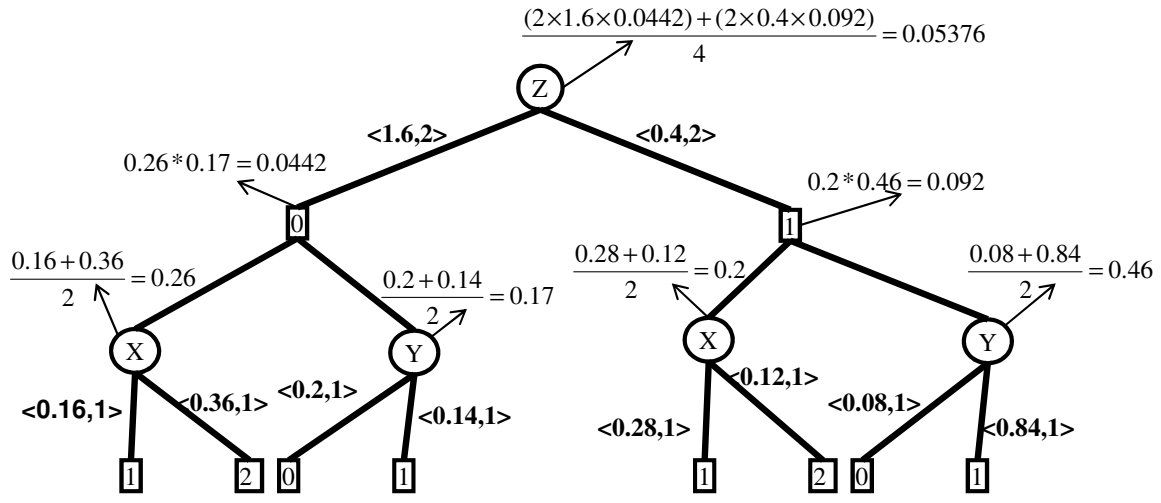

Figure 3: Computation of Values of OR and AND nodes in a AND/OR sample tree. The value of root node is equal to the AND/OR sample mean

Bayesian network given in Figure 2 are shown in Figure 3. Each AND node and OR node in Figure 3 is marked with a value that is computed recursively using definition 3.3. The value of OR nodes $X$ and $Y$ given $Z = j \in \{0,1\}$ is equal to $g_X\widehat{(Z=j)}$ and $g_Y\widehat{(Z=j)}$ respectively defined in Equation 9. The value of the root node is equal to the AND/OR sample mean which is equal to the sample mean computed by parts in Equation 10.

---

**Algorithm 1** AND/OR Importance Sampling

---

**Input:** an ordering $O = (X_1, \ldots, X_n)$,a Bayesian network BN and a proposal distribution $Q$
**Output:** Estimate of Probability of Evidence

1: Generate samples $\mathbf{x}^1, \ldots, \mathbf{x}^N$ from $Q$ along $O$.
2: Build a AND/OR sample tree $S_{AOT}$ for the samples $\mathbf{x}^1, \ldots, \mathbf{x}^N$ along the ordering $O$.
3: Initialize all labeling functions $\langle w, \# \rangle$ on each arc from an Or-node $n$ to an And-node $n'$ using Definition 3.1.
4: **FOR** all leaf nodes $i$ of $S_{AOT}$ do
5:   **IF And-node** v(i)= 1 **ELSE** v(i)=0
6: **For** every node $n$ from leaves to the root do
7:   Let $C(n)$ denote the child nodes of node $n$
8:   **IF** $n = \langle X, x \rangle$ is a AND node, then $v(n) = \prod_{n' \in C(n)} v(n')$
9:   **ELSE** if $n = X$ is a OR node then

$$v(n) = \frac{\sum_{n' \in C(n)} (\#(n,n') w(n,n') v(n'))}{\sum_{n' \in C(n)} \#(n,n')}.$$

10: Return v(root node)

---

We now have the necessary definitions to formally present the AND/OR importance sampling scheme (see Algorithm 1). In Steps 1-3, the algorithm generates samples from $Q$ and stores them on an AND/OR sample tree. The algorithm then computes the AND/OR sample mean over the AND/OR sample tree recursively from leaves to the root in Steps $4-9$. We can show that the value $v(n)$ of a node in the AND/OR sample tree stores the sample average of the subproblem rooted at $n$, subject to the current variable instantiation along the path from the root to $n$. If $n$ is the root, then $v(n)$ is the AND/OR sample mean which is our AND/OR estimator of probability of evidence. Finally, we summarize the complexity of computing AND/OR sample mean in the following theorem:

THEOREM **3.6.** *Given N samples and n variables (with constant domain size), the time complexity of computing AND/OR sample mean is $O(nN)$ (same as importance sampling) and its space complexity is $O(nN)$ (the space complexity of importance sampling is constant).*

# 4 Variance Reduction

In this section, we prove that the AND/OR sample mean may have lower variance than the sample mean computed using importance sampling (Equation 3).

THEOREM **4.1** (Variance Reduction)**.** *Variance of AND/OR sample mean is less than or equal to the variance of importance sampling sample mean.*

*Proof.* The details of the proof are quite complicated and therefore we only provide the intuitions involved. As noted earlier the guiding principle of AND/OR sample mean is to take advantage of conditional independence in the graphical model. Let us assume that we have three random variables $\mathbf{X}$, $\mathbf{Y}$ and $\mathbf{Z}$ with the following relationship: $\mathbf{X}$ and $\mathbf{Y}$ are independent of each other given $\mathbf{Z}$ (similar to our example Bayesian network). The expression for variance derived here can be used in an induction step (induction is carried on the nodes of the pseudo tree) to prove the theorem.

In this case, importance sampling generates samples $((\mathbf{x}^1, \mathbf{y}^1, \mathbf{z}^1), \ldots, (\mathbf{x}^N, \mathbf{y}^N, \mathbf{z}^N))$ along the order $\langle \mathbf{Z}, \mathbf{X}, \mathbf{Y} \rangle$ and estimates the mean as follows:

$$\mu^{IS}(\mathbf{XYZ}) = \frac{\sum_{i=1}^{N} \mathbf{x}^i \mathbf{y}^i \mathbf{z}^i}{N} \tag{12}$$

Without loss of generality, let $\{\mathbf{z_1}, \mathbf{z_2}\}$ be the domain of $\mathbf{Z}$ and let these values be sampled $N_1$ and $N_2$ times respectively. We can rewrite Equation 12 as follows:

$$\mu^{IS}(\mathbf{XYZ}) = \frac{1}{N} \sum_{j=1}^{2} N_j \mathbf{z_j} \frac{\sum_{i=1}^{N} \mathbf{x}^i \mathbf{y}^i I(\mathbf{z}_j, \mathbf{x}^i, \mathbf{y}^i)}{N_j} \tag{13}$$

where $I(\mathbf{z}_j, \mathbf{x}^i, \mathbf{y}^i)$ is an indicator function which is 1 iff the partial assignment $(\mathbf{z}_j, \mathbf{x}^i, \mathbf{y}^i)$ is generated in any of the $N$ samples and 0 otherwise.

AND/OR sample mean is defined as:

$$\mu^{AO}(\mathbf{XYZ}) = \frac{1}{N} \sum_{j=1}^{2} N_j \mathbf{z}_j \left( \frac{\sum_{i=1}^{N} \mathbf{x}^i I(\mathbf{z}_j, \mathbf{x}^i)}{N_j} \right) \left( \frac{\sum_{i=1}^{N} \mathbf{y}^i I(\mathbf{z}_j, \mathbf{y}^i)}{N_j} \right) \tag{14}$$

where $I(\mathbf{x}^j, \mathbf{z}_i)$ (and similarly $I(\mathbf{y}^j, \mathbf{z}_i)$) is an indicator function which equals 1 when one of the $N$ samples contains the tuple $(\mathbf{x}^j, \mathbf{z}_i)$ (and similarly $(\mathbf{y}^j, \mathbf{z}_i)$)) and is 0 otherwise.

By simple algebraic manipulations, we can prove that the variance of estimator $\mu^{IS}(\mathbf{XYZ})$ is given by:

$$Var(\mu^{IS}(\mathbf{XYZ})) = \Bigg( \sum_{j=1}^{2} \mathbf{z}_j^2 Q(\mathbf{z_j}) \Big( \mu(\mathbf{X}|\mathbf{z}_j)^2 V(\mathbf{Y}|\mathbf{z}_j) + \mu(\mathbf{Y}|\mathbf{z}_j)^2 V(\mathbf{X}|\mathbf{z}_j) + V(\mathbf{X}|\mathbf{z}_j) V(\mathbf{Y}|\mathbf{z_j}) \Big) \Bigg) / N - \mu_{\mathbf{XYZ}}^2 / N \tag{15}$$

Similarly, the variance of AND/OR sample mean is given by:

$$Var(\mu^{AO}(\mathbf{XYZ})) = \left( \sum_{j=1}^{2} \mathbf{z}_j^2 Q(\mathbf{z_j}) \Big( \mu(\mathbf{X}|\mathbf{z}_j)^2 V(\mathbf{Y}|\mathbf{z}_j) + \mu(\mathbf{Y}|\mathbf{z}_j)^2 V(\mathbf{X}|\mathbf{z}_j) + \frac{V(\mathbf{X}|\mathbf{z}_j)V(\mathbf{Y}|\mathbf{z_j})}{N_j} \Big) \right) / N - \mu^2_{\mathbf{XYZ}}/N \quad (16)$$

where $\mu(\mathbf{X}|\mathbf{z}_j)$ and $V(\mathbf{X}|\mathbf{z}_j)$ are the conditional mean and variance respectively of $\mathbf{X}$ given $\mathbf{Z} = \mathbf{z}_j$. Similarly, $\mu(\mathbf{Y}|\mathbf{z}_j)$ and $V(\mathbf{Y}|\mathbf{z}_j)$ are the conditional mean and variance respectively of $\mathbf{Y}$ given $\mathbf{Z} = \mathbf{z}_j$.

From Equations 15 and 16, if $N_j = 1$ for all $j$, then we can see that the $Var(\mu^{AO}(\mathbf{XYZ})) = Var(\mu^{IS}(\mathbf{XYZ}))$. However if $N_j > 1$, $Var(\mu^{AO}(\mathbf{XYZ})) < Var(\mu^{IS}(\mathbf{XYZ}))$. This proves that the variance of AND/OR sample mean is less than or equal to the variance of conventional sample mean on this special case. As noted earlier using this case in induction over the nodes of a general pseudo-tree completes the proof.

□

# 5 Estimation in AND/OR graphs

Next, we describe a more powerful algorithm for estimating mean in AND/OR-space by moving from AND/OR-trees to AND/OR graphs as presented in [Dechter and Mateescu, 2007]. An AND/OR-tree may contain nodes that root identical subtrees. When such unifiable nodes are merged, the tree becomes a graph and its size becomes smaller. Some unifiable nodes can be identified using contexts defined below.

**Definition 5.1 (Context).** Given a belief network and the corresponding AND/OR search tree $S_{AOT}$ relative to a pseudo-tree $T$, the context of any AND node $\langle X_i, x_i \rangle \in S_{AOT}$ , denoted by $context(X_i)$, is defined as the set of ancestors of $X_i$ in $T$, that are connected to $X_i$ and descendants of $X_i$.

The context minimal AND/OR graph is obtained by merging all the context unifiable AND nodes. The size of the largest context is bounded by the tree width $w^*$ of the pseudo-tree [Dechter and Mateescu, 2007]. Therefore, the time and space complexity of a search algorithm traversing the context-minimal AND/OR graph is $O(exp(w^*))$.

**Example 5.2.** For illustration, consider the context-minimal graph in Figure 1(e) of the pseudo-tree from Figure 1(c). Its size is far smaller that that of the AND/OR tree from Figure 2(c) (30 nodes vs. 38 nodes). The contexts of the nodes can be read from the pseudo-tree in Figure 1(b) as follows: context(A) = {A}, context(B) = {B,A}, context(C) = {C,B,A}, context(D) = {D,C,B} and context(E) = {E,A,B}.

The main idea in AND/OR-graph estimation is to store all samples on an AND/OR-graph instead of an AND/OR-tree. Similar to an AND/OR sample tree, we can define an identical notion of an AND/OR sample graph.

**Definition 5.3 ( Arc labeled AND/OR sample graph).** Given a complete AND/OR graph $\phi_G$ and a set of samples $\mathbf{S}$ , an AND/OR sample graph $S_{AOG}$ is obtained by removing all nodes and arcs not in $\mathbf{S}$ from $\phi_G$. The labels on $S_{AOG}$ are set similar to that of an AND/OR sample tree (see Definition 3.1).

**Example 5.4.** The bold edges and nodes in Figure 1(c) define an AND/OR sample tree. The bold edges and nodes in Figure 1(d) define an AND/OR sample graph corresponding to the same samples that define the AND/OR sample tree in Figure 1(c).

The algorithm for computing the sample mean on AND/OR sample graphs is identical to the algorithm for AND/OR-tree (Steps 4-10 of Algorithm 1). The main reason in moving from trees to graphs is that the variance of the sample mean computed on an AND/OR sample graph can be even smaller than that computed on an AND/OR sample tree. More formally,

THEOREM **5.5.** *Let $V(\mu_{AOG})$, $V(\mu_{AOT})$ and $V(\mu_{IS})$ be the variance of AND/OR sample mean on an AND/OR sample graph, variance of AND/OR sample mean on an AND/OR sample tree and variance of sample mean of importance sampling respectively. Then given the same set of input samples:*

$$V(\mu_{AOG}) \leq V(\mu_{AOT}) \leq V(\mu_{IS})$$

We omit the proof due to lack of space.

THEOREM **5.6** (Complexity of computing AND/OR graph sample mean)**.** *Given a graphical model with $n$ variables, a psuedo-tree with treewidth $w^*$ and $N$ samples, the time complexity of AND/OR graph sampling is $O(nNw^*)$ while its space complexity is $O(nN)$.*

# 6 Experimental Evaluation

## 6.1 Competing Algorithms

The performance of importance sampling based algorithms is highly dependent on the proposal distribution [Cheng and Druzdzel, 2000]. It was shown that computing the proposal distribution from the output of a Generalized Belief Propagation scheme of Iterative Join Graph Propagation (IJGP) yields better empirical performance than other available choices [Gogate and Dechter, 2005]. Therefore, we use the output of IJGP to compute the proposal distribution $Q$. The complexity of IJGP is time and space exponential in its $i$-bound, a parameter that bounds cluster sizes. We use a $i$-bound of 5 in all our experiments.

We experimented with three sampling algorithms for benchmarks which do not have determinism: (a) (pure) IJGP-sampling, (b) AND/OR-tree IJGP-sampling and (c) AND/OR-graph IJGP-sampling. Note that the underlying scheme for generating the samples is identical in all the methods. What changes is the method of accumulating the samples and deriving the estimates. On benchmarks which have zero probabilities or determinism, we use the SampleSearch scheme introduced by [Gogate and Dechter, 2007] to overcome the rejection problem. We experiment with the following versions of SampleSearch on deterministic networks: (a) pure SampleSearch, (b) AND/OR-tree SampleSearch and (c) AND/OR-graph SampleSearch.

### 6.1.1 Results

We experimented with three sets of benchmark belief networks (a) Random networks, (b) Linkage networks and (c) Grid networks. Note that only linkage and grid networks have zero probabilities on which we use SampleSearch.The exact $P(\mathbf{e})$ for most instances is available from the UAI 2006 competition web-site.

Our results are presented in Figures 4-6. Each Figure shows approximate probability of evidence as a function of time. The bold line in each Figure indicates the exact probability of evidence. The reader can visualize the error from the distance between the approximate curves and the exact line. For lack of space, we show only part of our results. Each Figure shows the number of variables $n$, the maximum-domain size $d$ and the number of evidence nodes $|E|$ for the respective benchmark.

**Random Networks** From Figures 4(a) and 4(b), we see that AND/OR-graph sampling is better than AND/OR-tree sampling which in turn is better than pure IJGP-sampling. However there is not much difference in the error because the proposal distribution seems to be a very good approximation of the posterior.

**Grid Networks** All Grid instances have 1444 binary nodes and between 5-10 evidence nodes. From Figures 5(a) and 5(b), we can see that AND/OR-graph SampleSearch and AND/OR-tree SampleSearch are substantially better than pure SampleSearch.

**Linkage Networks** The linkage instances are generated by converting a Pedigree to a Bayesian network [Fishelson and Geiger, 2003]. These networks have between 777-2315 nodes with a maximum domain size of 36. Note that it is hard to compute exact probability of evidence in these networks [Fishelson and Geiger, 2003]. We observe from Figures 6(a),(b) (c) and (d) that AND/OR-graph SampleSearch is substantially more accurate than AND/OR-tree SampleSearch which in turn is substantially more accurate than pure SampleSearch. Notice the log-scale in Figures 6 (a)-(d) which means that there is an order of magnitude difference between the errors. Our results suggest that AND/OR-graph and tree estimators yield far better performance than conventional estimators especially on problems in which the proposal distribution is a bad approximation of the posterior distribution.

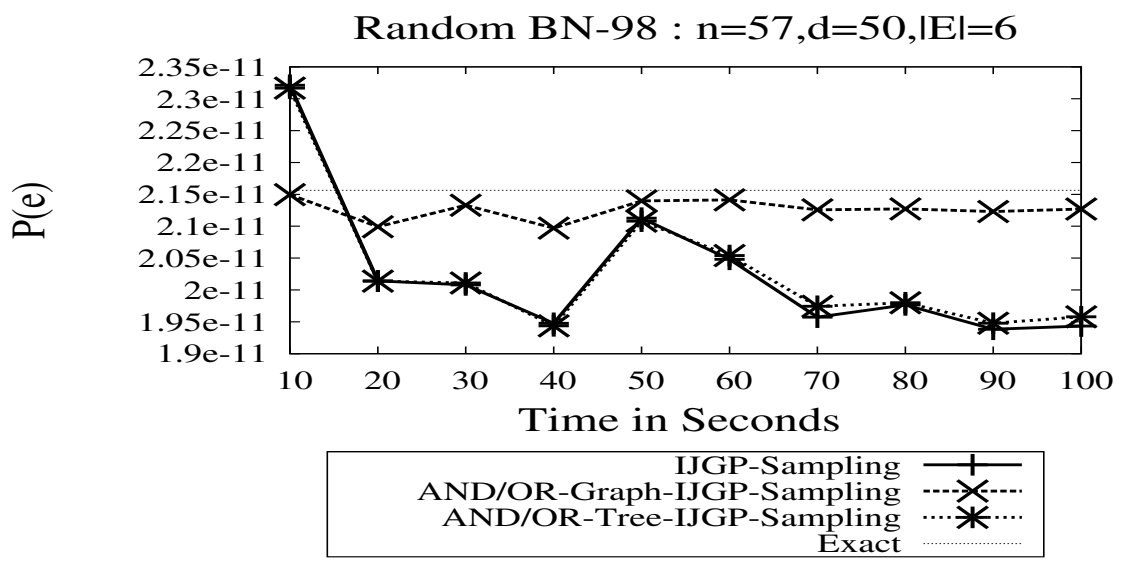

(a)

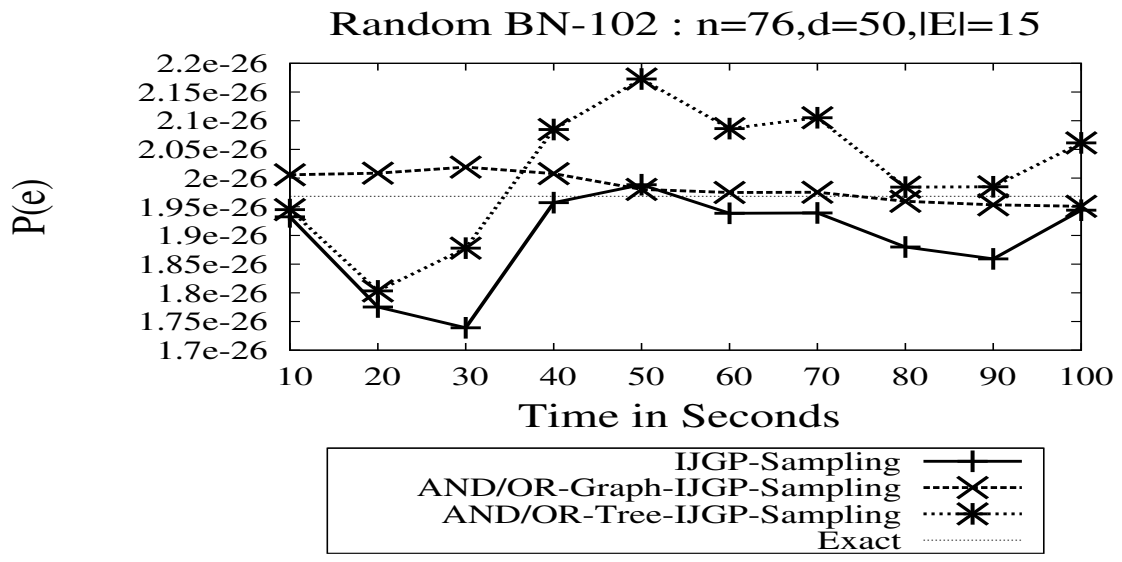

(b)

Figure 4: Random Networks

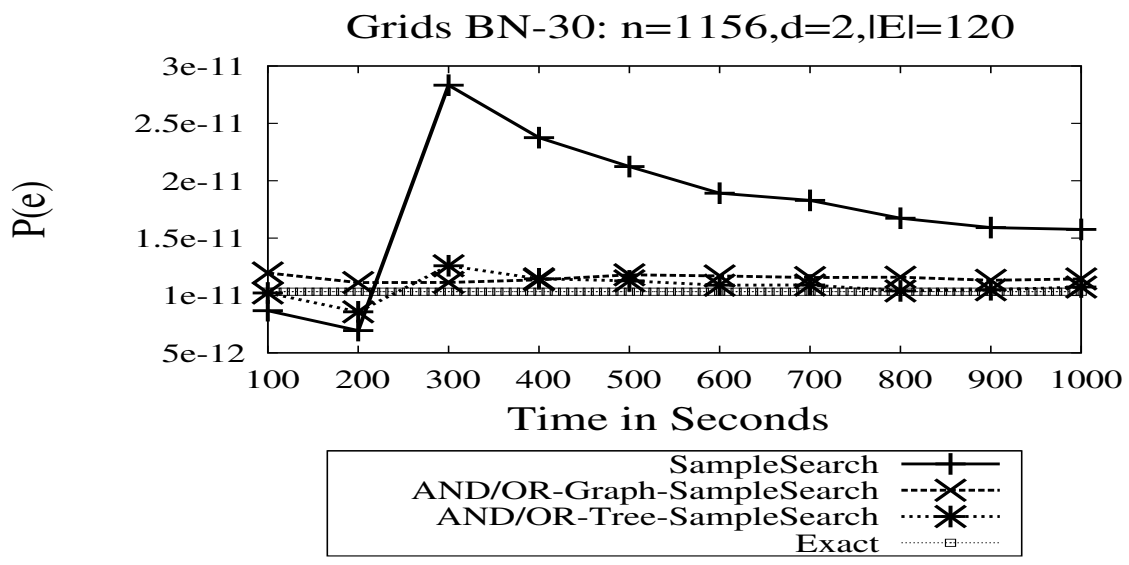

(a)

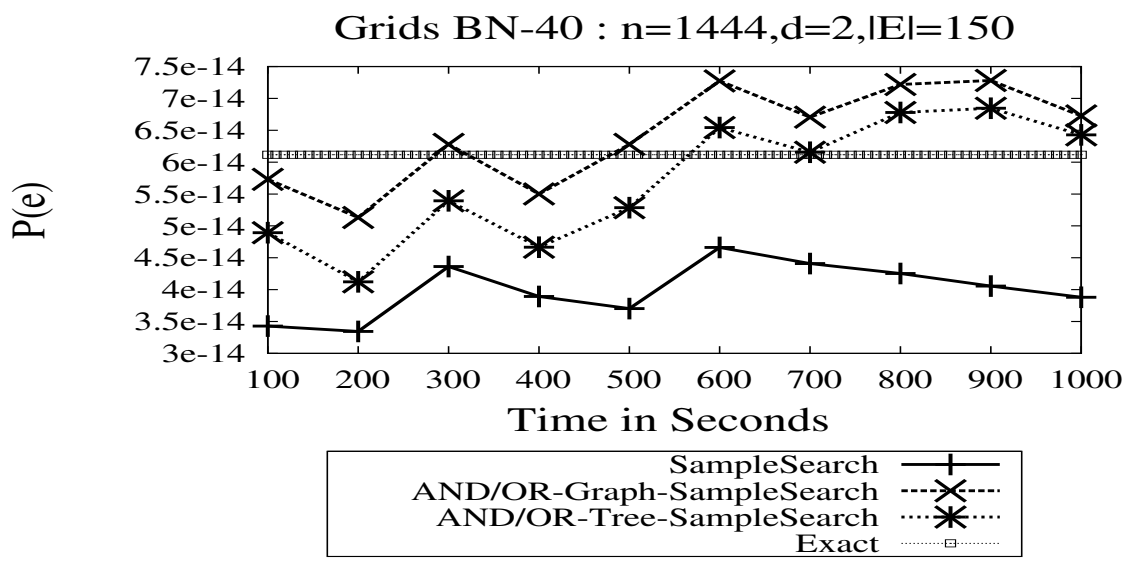

(b)

Figure 5: Grid Networks

## 7 Related Work and Summary

The work presented in this paper is related to the work by [Hernndez and Moral, 1995, Kjærulff, 1995, Dawid et al., 1994] who perform sampling based inference on a junction tree. The main idea in these papers is to perform message passing on a junction tree by substituting messages which are too hard to compute exactly by their sampling-based approx-imations. [Kjærulff, 1995, Dawid et al., 1994] use Gibbs sampling while [Hernndez and Moral, 1995] use importance sampling to approximate the messages. Similar to recent work on Rao-Blackwellised sampling such as [Bidyuk and Dechter, 2003, Paskin, 2004, Gogate and Dechter, 2005], variance reduction is achieved in these junction tree based sampling schemes because of some exact computations; as dictated by the Rao-Blackwell theorem. AND/OR estimation, however, does not require exact computations to achieve variance reduction. In fact, variance reduction due to Rao-Blackwellisation is orthogonal to the variance reduction achieved by AND/OR estimation and therefore the two could be combined to achieve more variance reduction. Also, unlike our work which focuses on probability of evidence, the focus of these aforementioned papers was on belief updating.

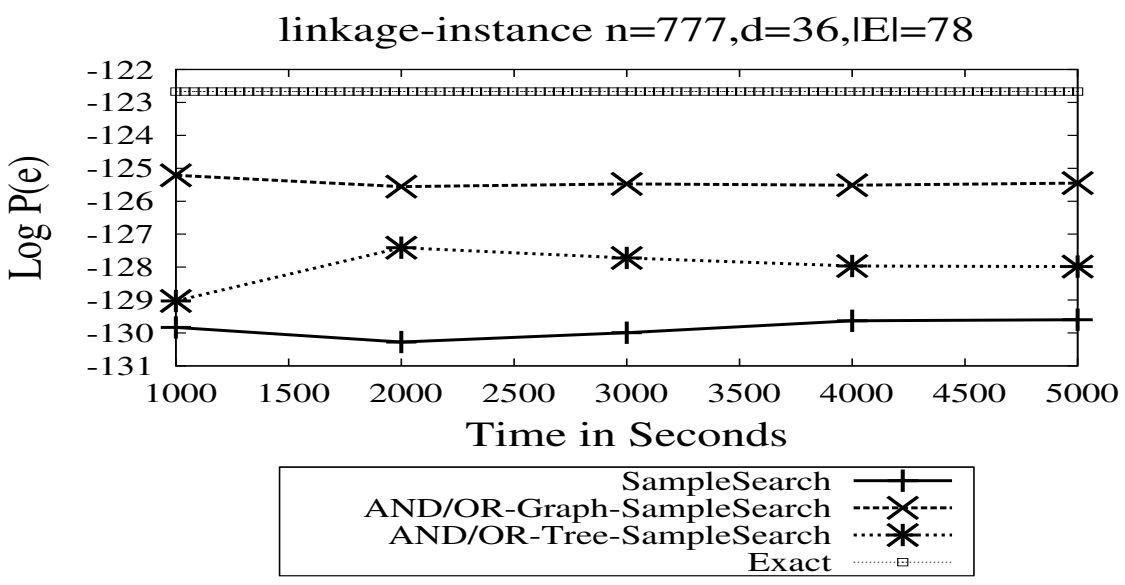

(a)

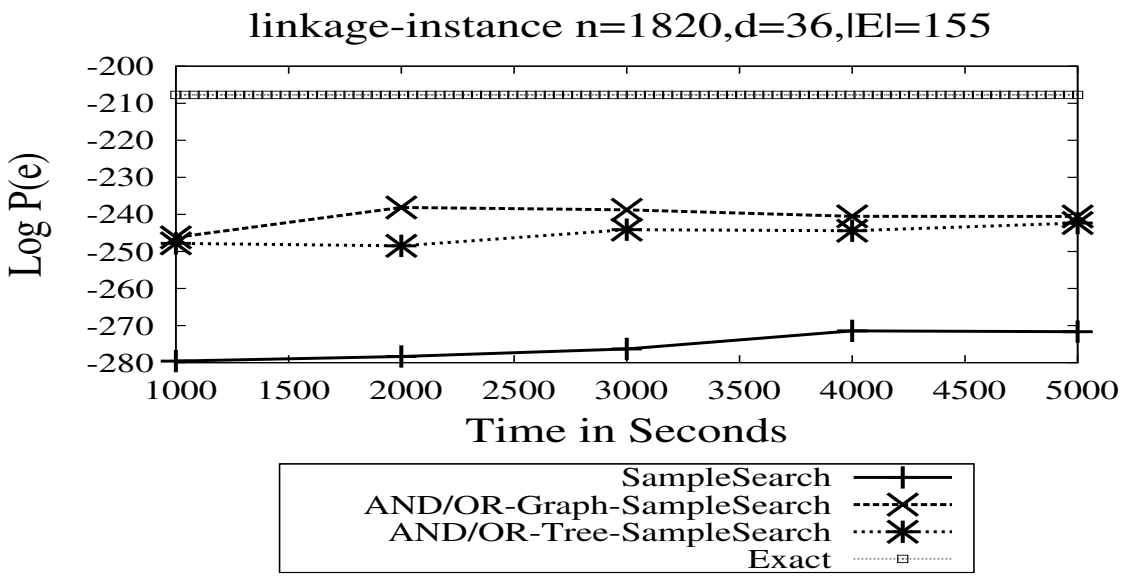

(b)

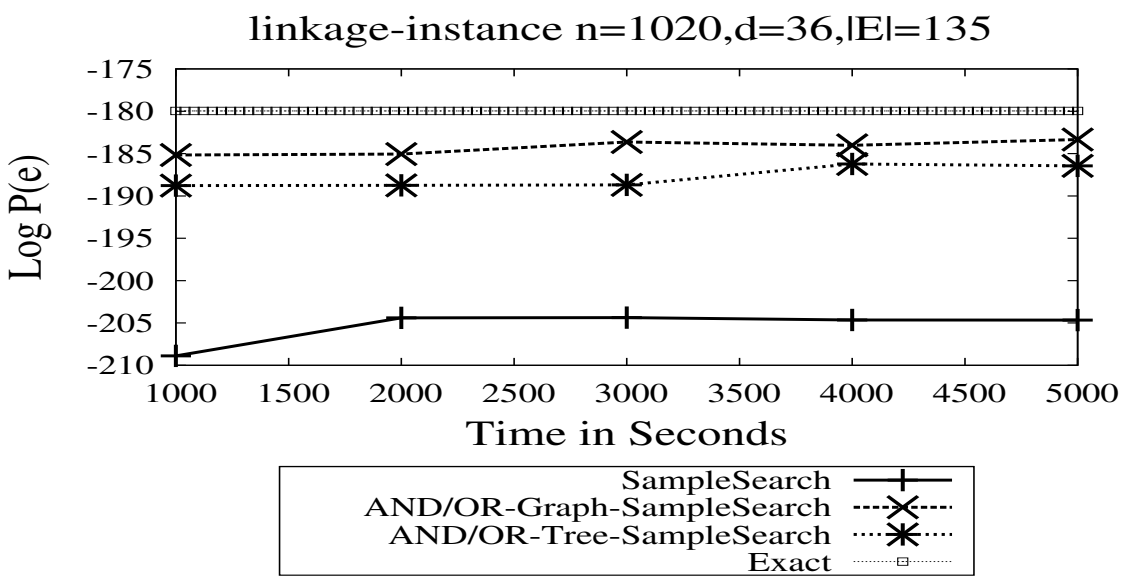

(c)

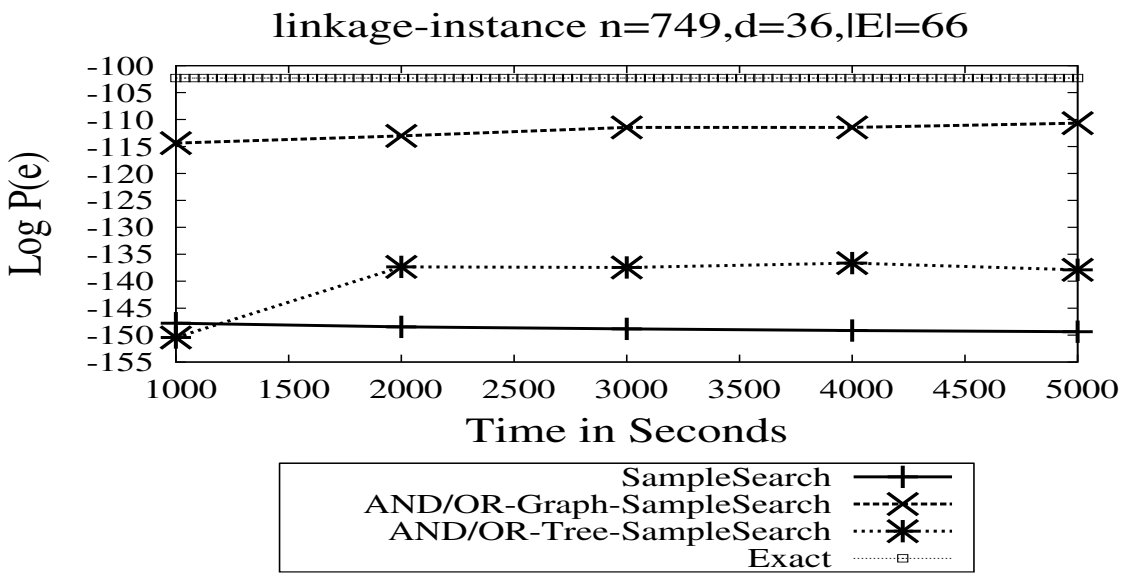

(d)

Figure 6: Linkage Bayesian Networks

To summarize, the paper introduces a new sampling based estimation technique called AND/OR importance sampling. The main idea of our new scheme is to derive statistics on the generated samples by using an AND/OR tree or graph that takes advantage of the independencies present in the graphical model. We proved that the sample mean computed on an AND/OR tree or graph may have smaller variance than the sample mean computed using the conventional approach. Our experimental evaluation is preliminary but quite promising showing that on most instances AND/OR sample mean has lower error than importance sampling and sometimes by significant margins.

## Acknowledgements

This work was supported in part by the NSF under award numbers IIS-0331707, IIS-0412854 and IIS-0713118 and the NIH grant R01-HG004175-02.